\documentclass[sigconf]{acmart}
\usepackage{dsfont}
\usepackage{caption}
\usepackage{subcaption}

\AtBeginDocument{%
  \providecommand\BibTeX{{%
    \normalfont B\kern-0.5em{\scshape i\kern-0.25em b}\kern-0.8em\TeX}}}

\copyrightyear{2021} 
\acmYear{2021} 
\setcopyright{acmcopyright}\acmConference[ICMI '21]{Proceedings of the 2021 International Conference on Multimodal Interaction}{October 18--22, 2021}{Montréal, QC, Canada}
\acmBooktitle{Proceedings of the 2021 International Conference on Multimodal Interaction (ICMI '21), October 18--22, 2021, Montréal, QC, Canada}
\acmPrice{15.00}
\acmDOI{10.1145/3462244.3479955}
\acmISBN{978-1-4503-8481-0/21/10}

\begin{document}


\title{Self-supervised Contrastive Learning of Multi-view Facial Expressions}


\author{Shuvendu Roy, Ali Etemad}
\affiliation{%
  \institution{Department of Electrical and Computer Engineering \& Ingenuity Labs Research Institute\\
    Queen's University}
  \city{Kingston}
  \country{Canada}
}
\email{{shuvendu.roy, ali.etemad}@queensu.ca}


\begin{abstract}
Facial expression recognition (FER) has emerged as an important component of human-computer interaction systems. Despite recent advancements in FER, performance often drops significantly for non-frontal facial images. We propose Contrastive Learning of Multi-view facial Expressions (CL-MEx) to exploit facial images captured simultaneously from different angles towards FER. CL-MEx is a two-step training framework. In the first step, an encoder network is pre-trained with the proposed self-supervised contrastive loss, where it learns to generate view-invariant embeddings for different views of a subject. The model is then fine-tuned with labeled data in a supervised setting. We demonstrate the performance of the proposed method on two multi-view FER datasets, KDEF and DDCF, where state-of-the-art performances are achieved. Further experiments show the robustness of our method in dealing with challenging angles and reduced amounts of labeled data.
\end{abstract}



\maketitle

\section{Introduction}
Facial expression recognition (FER) is an important attribute of non-verbal communication. FER systems play an important role in human-machine systems by enabling customization and adaptation of such systems to user reactions. Examples of such devices include emotion-aware multimedia and smart devices \cite{cho2019instant}, personal mood management systems \cite{thrasher2011mood, sanchez2013inferring}, driving assistants \cite{leng2007experimental}, health-care assistants \cite{tokuno2011usage}, and others.
However, despite the potential for a wide variety of applications, FER remains challenging due to various factors like illuminations, scene backgrounds, occlusions, and challenging viewing angles among others. 

Despite the recent progress in FER systems with deep learning \cite{MPCNN, PhaNet, ssl_affectnet}, most prior work in the area focus on only utilizing frontal facial views \cite{ssl_affectnet, breg_next}. As a result, developed models may not be able to recognize expressions in faces captured from sharp side angles. Another problem in deep learning-based FER systems is the requirement for large amounts of training data given the millions of trainable parameters.

\begin{figure}[t]
  \centering
  \includegraphics[width=0.65\linewidth]{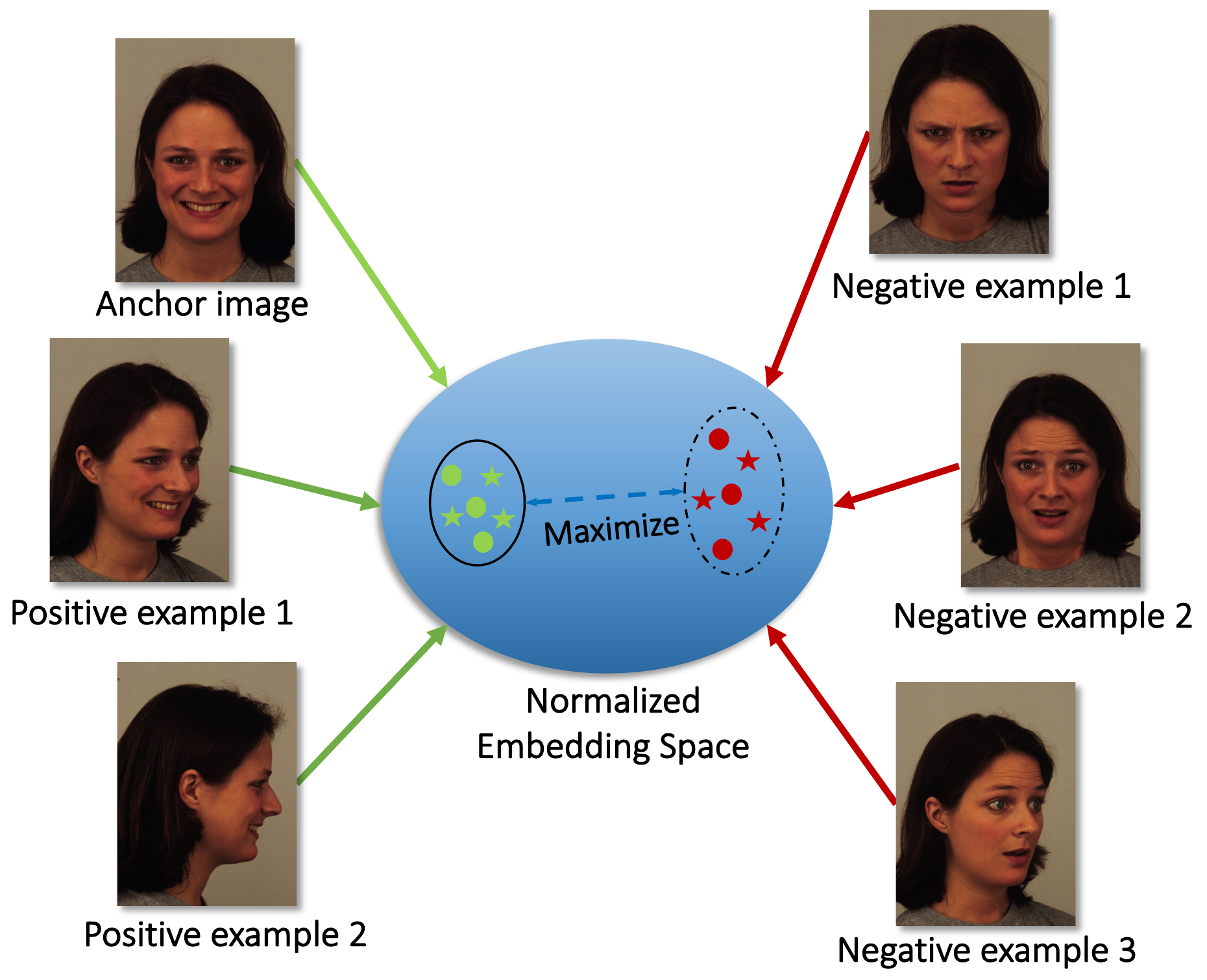}
  \caption{High-level visualization of CL-MEx. The method aims to minimize the distance between all positive examples (i.e. the radius of the black ellipses) which include similar expressions obtained from different angles, and maximize the distance between positive and negative pairs. Subject ID of the depicted face: AF01.}
  \label{fig:ssl}
\end{figure}

In this work, to deal with the problems of view-point sensitivity and labeled training set size, we propose \textbf{C}ontrastive \textbf{L}earning for \textbf{M}ulti-view facial \textbf{Ex}pressions (CL-MEx). Our proposed framework learns facial expression representations through a two-step training process. The first step is self-supervised pre-training, whereby our novel loss function forces the latent representations of `different' views with the same expression from the same subject to be grouped together. This will allow for challenging viewpoints (e.g. side view) to be augmented by and benefit from the more effective representations of frontal views, allowing better performance (see Figure \ref{fig:ssl}). After pre-training, we fine-tune the model with the labeled data in a supervised setting. Our experiments show that the CL-MEx training protocol obtains better results than fully supervised training and outperforms the state-of-the-art solutions on the two datasets used (KDEF \cite{kdef} and DDCF \cite{ddcf}). Moreover, we demonstrate that our model does not require all the available labels to reach the performance of a fully supervised model. Lastly, we demonstrate that as a result of our proposed loss, our model is better equipped to deal with challenging views.

In summary, our contributions are as follows.
\begin{itemize}
\item We propose a novel contrastive learning framework, CL-MEx, for multi-view data. Our model utilizes a novel loss function designed specifically for multi-view images.
 
\item Experiments demonstrate the strong performance of our approach by setting a new \textit{state-of-the-art}, and reduced sensitivity to challenging views. Our model also requires less labeled data to obtain competitive results with respect to fully supervised training. 

\end{itemize}

\section{Related Works}
In this section, we review recent works in two key areas related to this paper: self-supervised and contrastive learning, and multi-view facial expression recognition.

\subsection{Self-Supervised and Contrastive Learning}

To reduce the influence of output labels and human-annotations, self-supervised learning has been proposed to use specific augmentations \cite{chen2020simple} or transformations \cite{noroozi2017representation, dosovitskiy2015discriminative} applied to input data, and produce \textit{pseudo-labels} to train or pre-train the model. Successive to training or pre-training the model with the augmented/transformed data and associated pseudo-labels (`pretext' tasks), the main building blocks of the model, for instance the convolutional blocks in a CNN, can then be used for `downstream' tasks, e.g. classification \cite{dosovitskiy2015discriminative}. 
In this phase, the convolutional blocks are typically frozen and new fully connected layers are added and trained from scratch using the original output labels \cite{chen2020simple}. 
The pre-trained components can also be fine-tuned to help achieve better performance \cite{noroozi2016unsupervised}. 

Self-supervised learning has been utilized in a variety of different problem domains including image analysis \cite{noroozi2016unsupervised, chen2019self}, wearable-based activity recognition \cite{saeed2019multi, rahimi2020self}, and affective computing with bio-signals \cite{sarkar2020self, sarkar2020detection, sarkar2020}.
Specifically for image-based recognition, a variety of different self-supervised solutions have been proposed in recent years \cite{doersch2015unsupervised, noroozi2016unsupervised,zhang2016colorful}. A key differentiator in many of these methods is the design of novel and interesting pretext tasks. Examples include jigsaw puzzle solving (where the images is divided into pieces and shuffled with the pretext aiming to solve the puzzle) \cite{doersch2015unsupervised, noroozi2016unsupervised}, inpainting (where patches are removed and the pretext aims to complete the image) \cite{pathak2016context}, transformation classification (where the pretext aims to classify the type of transform applied to the image, e.g., rotation, mirroring, etc.) \cite{gidaris2018unsupervised}, and colorization (where the pre-text aims to color the image converted to gray-scale) \cite{zhang2016colorful}.

Contrastive learning is a variation of self-supervised learning that has recently shown great success, achieving state-of-the-art performance in a variety of tasks such as object detection \cite{yang2021instance}, medical image analysis \cite{chaitanya2020contrastive}, image classification \cite{hadsell2006dimensionality, dosovitskiy2014discriminative, oord2018representation, bachman2019learning}, and video-based affect classification \cite{roy2021}. A recent method called SimCLR \cite{simclr} proposed the use of transformations, referred to as `augmentations', to create different variations of input images and maximize the agreement between positive pairs (augmented versions of the same image) via contrastive loss. A number of different variations of SimCLR have been proposed to exploit this framework and provide robust results \cite{chen2020exploring, supcon, chen2020big}. Later, a variation of contrastive learning was proposed to also make use of the output labels, effectively creating a `supervised' contrastive learning framework \cite{supcon}. In this approach, the positive examples were generated not only from different augmentations of the same input, but also the augmented versions of other examples in the `same class'. 

In the context of FER, which is the focus of this paper, self-supervised or contrastive learning have rarely been explored. In a recent work, FER was performed with a combination of contrastive learning and rotation prediction as pretext tasks \cite{ssl_affectnet}.

\subsection{Multi-view FER}

Multi-view FER exploits facial images obtained from different angles to learn a mapping between images of the face performing specific types of expressions and the expression class. This approach allows for the methods to capture information that might otherwise be hidden from the camera in certain view-points, resulting in more robust performance. Deep learning solutions have recently been proposed to jointly learn the different  view-points and perform FER in a variety of different settings \cite{RB-FNN, MPCNN, PhaNet}.

A multi-channel pose-aware CNN (MPCNN) was proposed in \cite{MPCNN}, and showed strong performance for multi-view FER. In PhaNet \cite{PhaNet}, a pose-adaptive hierarchical attention network was proposed that could jointly recognize the facial expressions and poses. At any angle of view, the attention module helped the model identify the most relevant regions for detecting the expression. In \cite{sepas2021multi}, a new LSTM cell architecture was proposed to jointly learn different perspectives, which was then used for a number of different tasks, including multi-view FER.


\section{Method}

We propose a framework called CL-MEx for exploiting images acquired from multiple views to perform FER with self-supervised contrastive learning. Like SimCLR \cite{simclr}, our proposed method uses positive and negative examples to carry out contrastive learning. However, we do not limit the definition of positive pairs as augmentations of a particular input, and expand this notion to include the augmentations of all the views of the input image. CL-MEx then uses our proposed loss function to bring the representations of different views of a subject with the same expression closer in the embedding space (as shown earlier in Figure \ref{fig:ssl}). Successive to the pre-training step, we fine-tune the model by supervised learning using the labels from the dataset. The proposed method uses the multiple views of an image in the self-supervised pre-training step only, while at inference time, it predicts the expression using only a single-view image. Following we present the details of our proposed method including the encoder architecture, proposed loss, and implementation details.

\subsection{CL-MEx Components}
The main components of our proposed framework are as follows.


\noindent \textbf{Data Augmentation Module,} $Aug(.)$. This module randomly applies combinations of the following augmentations: random cropping, random resizing, random horizontal flip, random color distortion, and random gray-scaling, as $x' = Aug(x)$, where $x$ is the input image. The module applies random flip and gray-scaling with a probability of 0.5. For resize followed by random crop, first the image is randomly resized by a factor of 0.2 to 1, and then randomly cropped to obtain an augmented image of the original size. For colour distortion, the brightness, hue, saturation, and contrast are randomly changed with a coefficient of 0.4 to 1. For each input image we generate two augmented versions of the same image.

\noindent \textbf{Encoder Network,} $Enc(.)$, which maps $x'$ to a representation embedding vector, $r = Enc(x')$. We pass both of the augmented images to the same encoder, resulting in a pair of representation vectors. We use the embedding dimension of 512 for all our experiments. In this work, we use the ResNet-18 \cite{resnet} as our encoder network. 
    
\noindent \textbf{Projection Head,} $Proj(.)$, which maps the vector $r$ onto a projection embedding vector, $z = Proj(r)$. Following the intuition of \cite{simclr, supcon}, we use a 2-layer dense network for the projection head which gives an output dimension of 128. We normalize the output of this layer to enable using a linear product to measure distances in the projection space. This projection has proven to be important for learning effective representations by self-supervised pre-training \cite{simclr}. After the pre-training, we discard the projection head and only fine-tune the encoder for the FER task.

\begin{figure}[t]
  \centering
  \includegraphics[width=0.6\linewidth]{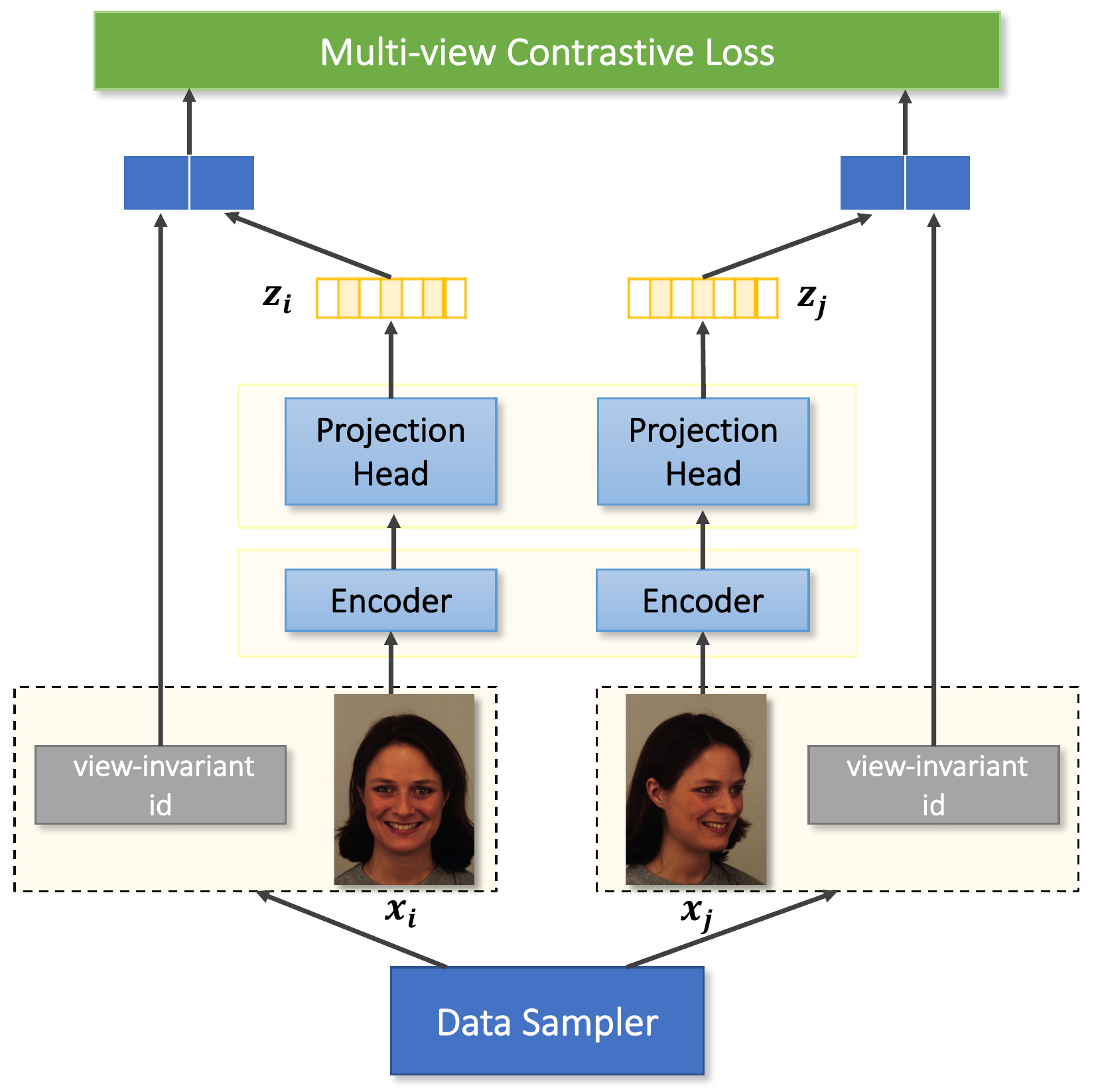}
  \caption{Training pipeline of the proposed method.}
  \label{fig:model}
\end{figure}

\subsection{Multi-view Contrastive Loss}
For the purpose of pre-training the model in a self-supervised setting, we propose a contrastive loss function which is inspired by SupCon \cite{supcon}. However, unlike SupCon, we only need the multi-view information to train the model and do not use the output labels in the pre-training stage. 

In a batch of size $2N$, the images are obtained by applying two random augmentations on $N$ random input samples. These images are passed through the encoder and the projection head to generate the embedding vector $Z$, on which the contrastive loss is calculated. The multi-view contrastive loss is defined as:
\begin{equation}
\small
\begin{split}
    \mathcal{L}^{\text{CL-MEx}} = \sum_{i=1}^{2N} \bigg( \frac{-1}{2N_{v_i}-1} \sum_{j=1}^{2N}
    \mathds{1}_{i\neq j} \cdot \mathds{1}_{v_i=v_j}
    \cdot \\
    log\frac{exp(z_i\cdot z_j/\tau)}{\sum_{k=1}^{2N} \mathds{1}_{[k \neq i]} exp(z_i \cdot z_k/\tau) } \bigg)~,
\end{split}
\end{equation}
where $v_i$ is the view-invariant id of image $i$, which indicates that for $v_i=v_j$, $i$ and $j$ are different views of the same subject. $\mathds{1}_{[a=b]} \in \{0, 1\}$ is a function of $a$ and $b$, which outputs 1 when $a=b$, and 0 otherwise. $N_{v_i}$ is the number of instances with id $v_i$. $\tau$ represents a temperature parameter which scales the dot product similarity. Figure \ref{fig:model} depicts a visual illustration of the proposed CL-MEx method, where the data sampler generates two augmented images along with their corresponding view-invariant ids, which generate the embedding by means of the proposed loss function.

\subsection{Downstream FER}
While the projection head is an important component for learning representations in the contrastive pre-training step, it is no longer required for the final downstream FER classification task. During the second step for FER, we discard the projection head and add a linear layer for predicting the output class probability. We then freeze the encoders and train the new linear layer with class labels using categorical cross-entropy loss. This is followed by fine-tuning the entire model including the encoders to maximize performance.

\subsection{Implementation Details}
During the pre-training step, the model is trained with the proposed multi-view contrastive loss function for 500 epochs with an Adam optimizer, a learning rate of 0.0001 with cosine learning rate decay, and weight decay of 1e-4. The downstream FER step consists of a total of 60 training epochs. First, we train the newly added classification layer for 10 iterations, and then fine-tune the full model for 50 additional iterations. During this step, an Adam optimizer was again used, this time with a plateau learning rate decay with an initial learning rate of 1e-4, decay factor of 0.5, and patience of 3. The method is implemented with PyTorch and trained on 4 NVIDIA V100 GPUs. All image resolutions are $224 \times 224$.

\section{Experiments and Results}

This section describes the experimental setup and results for CL-MEx on two benchmark datasets that contain multi-view facial expressions, Karolinska Directed Emotional Faces (KDEF) \cite{kdef} and Dartmouth Database of Children's Faces (DDCF) \cite{ddcf}.

\subsection{Datasets}

\textbf{KDEF} \cite{kdef} is a multi-view emotion recognition dataset with 7 expressions (afraid, angry, happy, sad, surprised, disgust, and neutral), where each subject is imaged from 5 different views (-90$^{\circ}$: full left (FL), -45$^{\circ}$: half left (HL), 0$^{\circ}$: straight (S), +45$^{\circ}$: half right (HR), and +90$^{\circ}$: full right (FR)). The dataset was collected from 140 subjects. 
The \textbf{DDCF} \cite{ddcf} is also a multi-view dataset collected from 80 subjects. It contains 8 expression classes (afraid, angry, happy, sad, surprised, disgust, pleased, and neutral), from 5 different views (-60$^{\circ}$: full left (FL), -30$^{\circ}$: half left (HL), 0$^{\circ}$: straight (S), +30$^{\circ}$: half right (HR), and +60$^{\circ}$: full right (FR)).

\begin{table}[!t]
    \caption{Results and comparison with other methods.}
    \begin{subtable}{.5\linewidth}
      \centering
        \caption{KDEF}
        
        \begin{tabular}{l l}
        \hline
         \textbf{Method} & \textbf{Acc. $\pm$ SD}\\
        \hline\hline
        SVM \cite{SVM} & 70.5$\pm$1.2\\
        SURF \cite{SURF} & 74.05$\pm$0.9\\
        TLCNN \cite{TLCNN} & 86.43$\pm$1.0\\
        PhaNet \cite{PhaNet} & 86.5\\
         MPCNN \cite{MPCNN} & 86.9$\pm$0.6\\
        RBFNN \cite {RB-FNN} & 88.87\\
        \textbf{CL-MEx} & \textbf{94.64$\pm$0.92}\\
        \hline
        \end{tabular}

    \end{subtable}%
    \begin{subtable}{.5\linewidth}
      \centering
        \caption{DDCF}
        
        \begin{tabular}{l l}
        \hline
         \textbf{Method} & \textbf{Acc. $\pm$ SD}\\
        \hline\hline

        LBP \cite {khan2019saliency} &  82.3\\
        SVM \cite {albu2015neural}& 91.27 \\
        RBFNN \cite{albu2015neural} & 91.80\\
        \textbf{CL-MEx} & \textbf{95.26$\pm$0.84}\\
        \hline
        \end{tabular}
        
    \end{subtable} 
    \label{tab:sota}
\end{table}

\begin{figure}[!t]
    \centering
    \begin{subfigure}[t]{0.2\textwidth}
        \centering
        \includegraphics[height=1.2in]{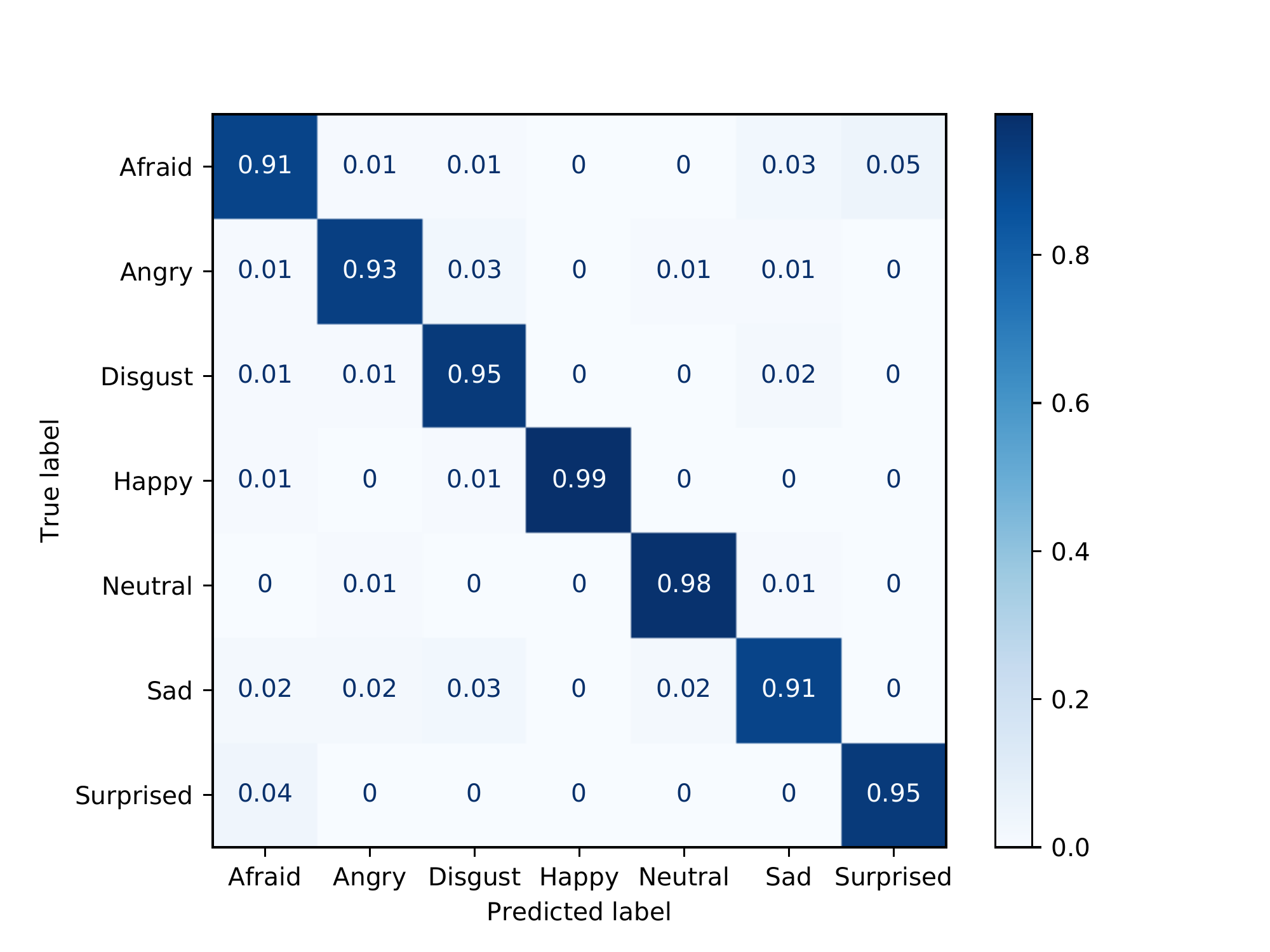}
        \caption{KDEF}
    \end{subfigure}
    ~
    \begin{subfigure}[t]{0.2\textwidth}
        \centering
        \includegraphics[height=1.2in]{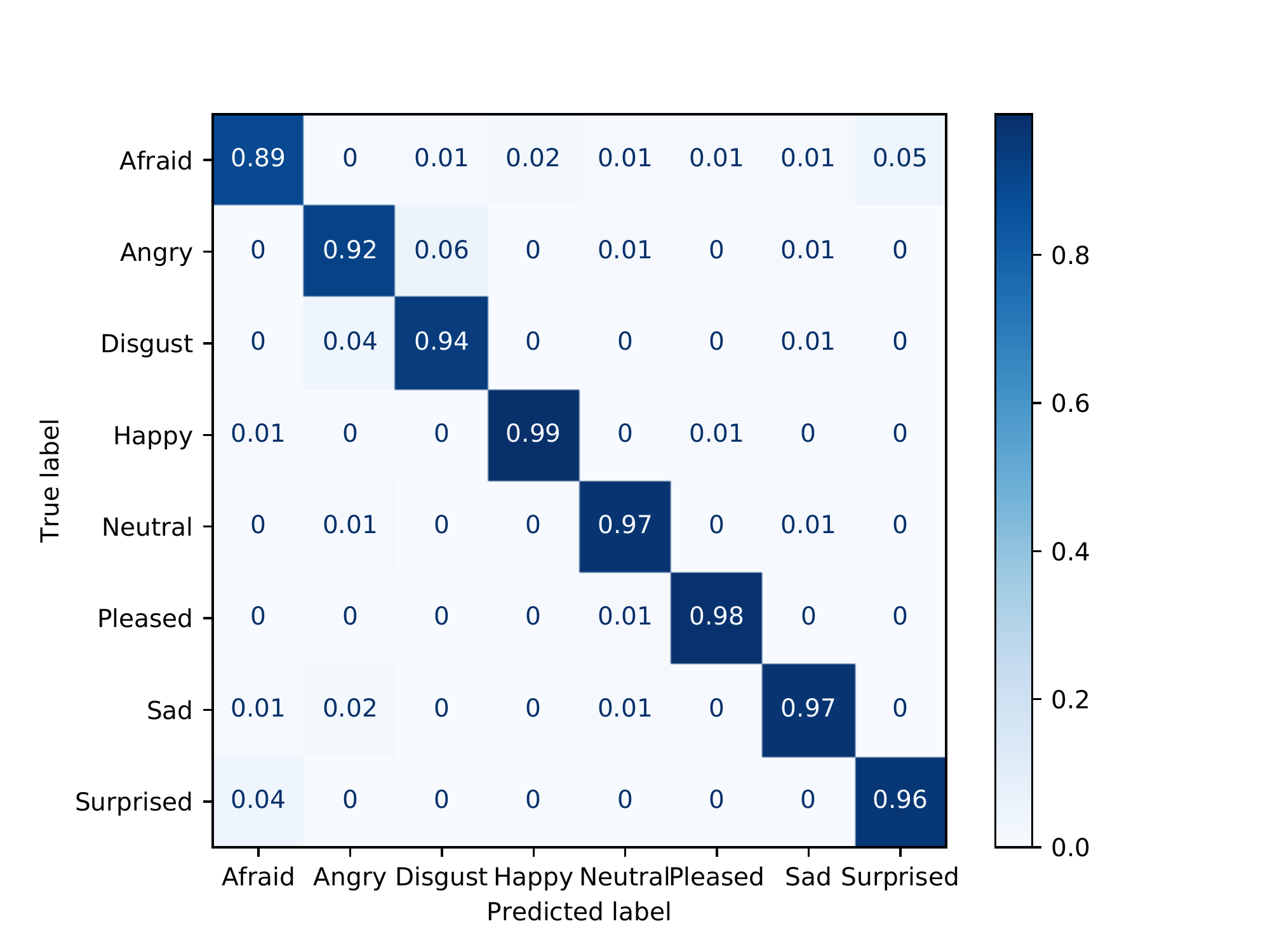}
        \caption{DDCF}
    \end{subfigure}
    \caption{Confusion matrices for different expressions.}
    \label{fig:confusion}
\end{figure}

\subsection{Results}

Accuracy is measured for the downstream network which predicts the expression labels. The input to this network is a single image (a single view of a subject). The accuracy is therefore measured as the rate of correct classifications over all the test images in 10-fold cross-validation.
Table \ref{tab:sota} compares the results of our proposed method with previous methods on KDEF and DDCF datasets. We observe that CL-MEx outperforms other methods on these two datasets when a single-view image is used during inference. It should be noted that some of the prior work mentioned in the table utilize all existing views to obtain the prediction.
Figure \ref{fig:confusion}(a) shows the confusion matrix for KDEF, where we observe that our method performs best on classifying `Happy' and `Neutral', while the least effective performance is obtained on `Afraid' and `Sad'. `Afraid' is often confused with `Sad' and `Surprised', whereas `Sad' is mostly confused with `Disgust'. In the case of the DDCF dataset, in Figure \ref{fig:confusion}(b) shows that strong performance is obtained for `Happy', `Neutral', `Pleased', and `Sad', while the method exhibits the least effectiveness on `Afraid' and `Angry', which are confused with `Surprised' and `Disgust' respectively. 

\begin{figure}[!t]
  \centering
  \includegraphics[width=0.7\linewidth]{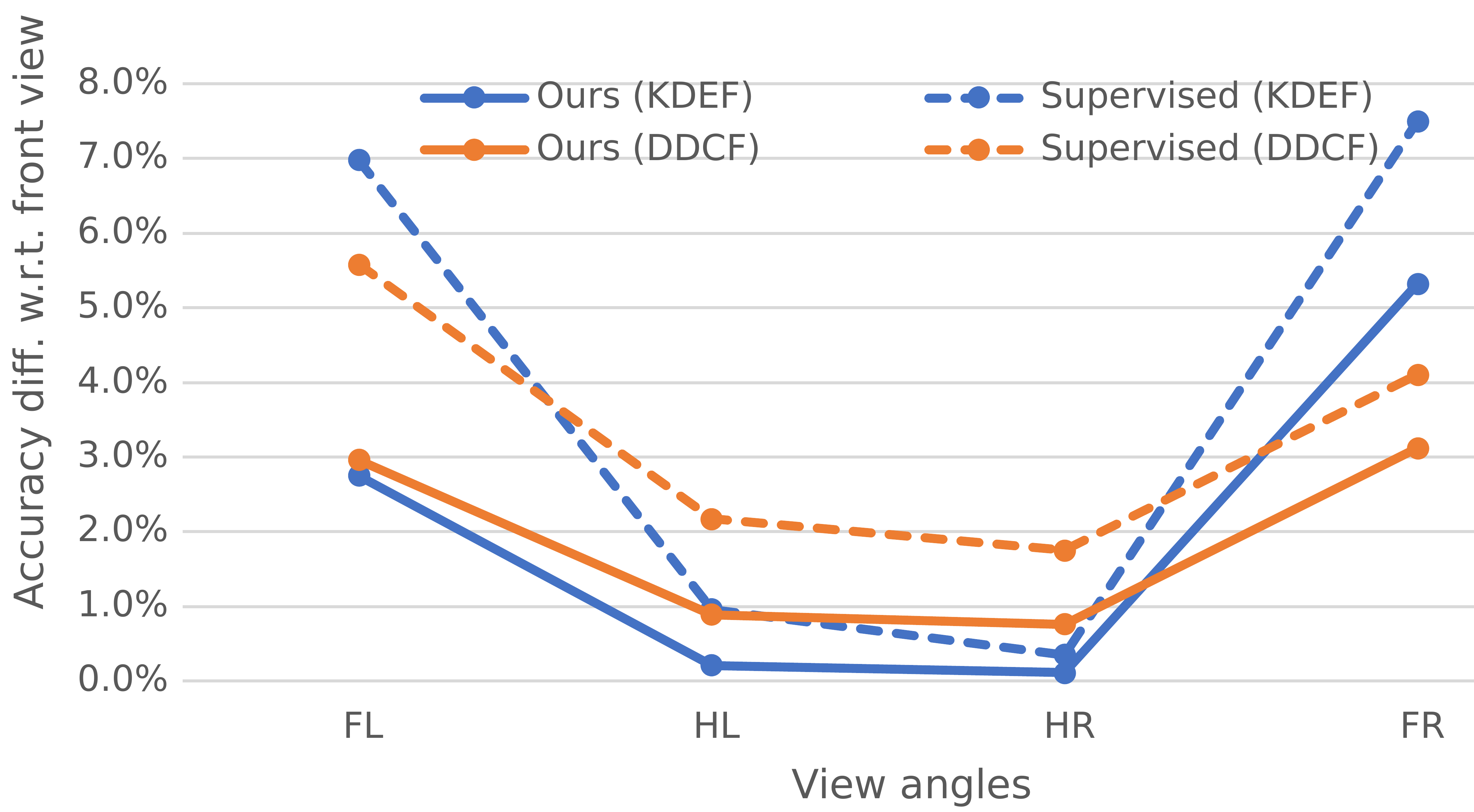}
  \caption{Drop in performance for different viewing angles with respect to the frontal view, for self-supervised CL-MEx and its fully supervised counterpart.}
  \label{fig:acc_diff}
\end{figure}

One of the main contributions of our proposed method is that it is robust to changes in viewing angles by forcing the representations obtained from different views to come together in the embedding space.
To objectively validate this notion, we compare the deterioration of the accuracy of the side-views with respect to the frontal view for CL-MEx in comparison to a fully supervised version of the same network (the model is trained with all the data and associated labels without the contrastive learning step). The results of this experiment are presented in Figure \ref{fig:acc_diff}, where we observe that across all side-view angles, our method shows a smaller drop in performance compared to supervised learning. 
It can be seen that as the angles become more severe, the supervised approach exhibits a larger drop, while our method shows less sensitivity. For example, in FL, the supervised method on KDEF has 7\% lower accuracy compared to its frontal view, whereas our method reduces that difference to only 2.8\%. In case of the less challenging views such as HL and HR, our method shows less than 1\% deviation for both of the datasets.

\begin{figure}[!t]
  \centering
  \includegraphics[width=0.7\linewidth]{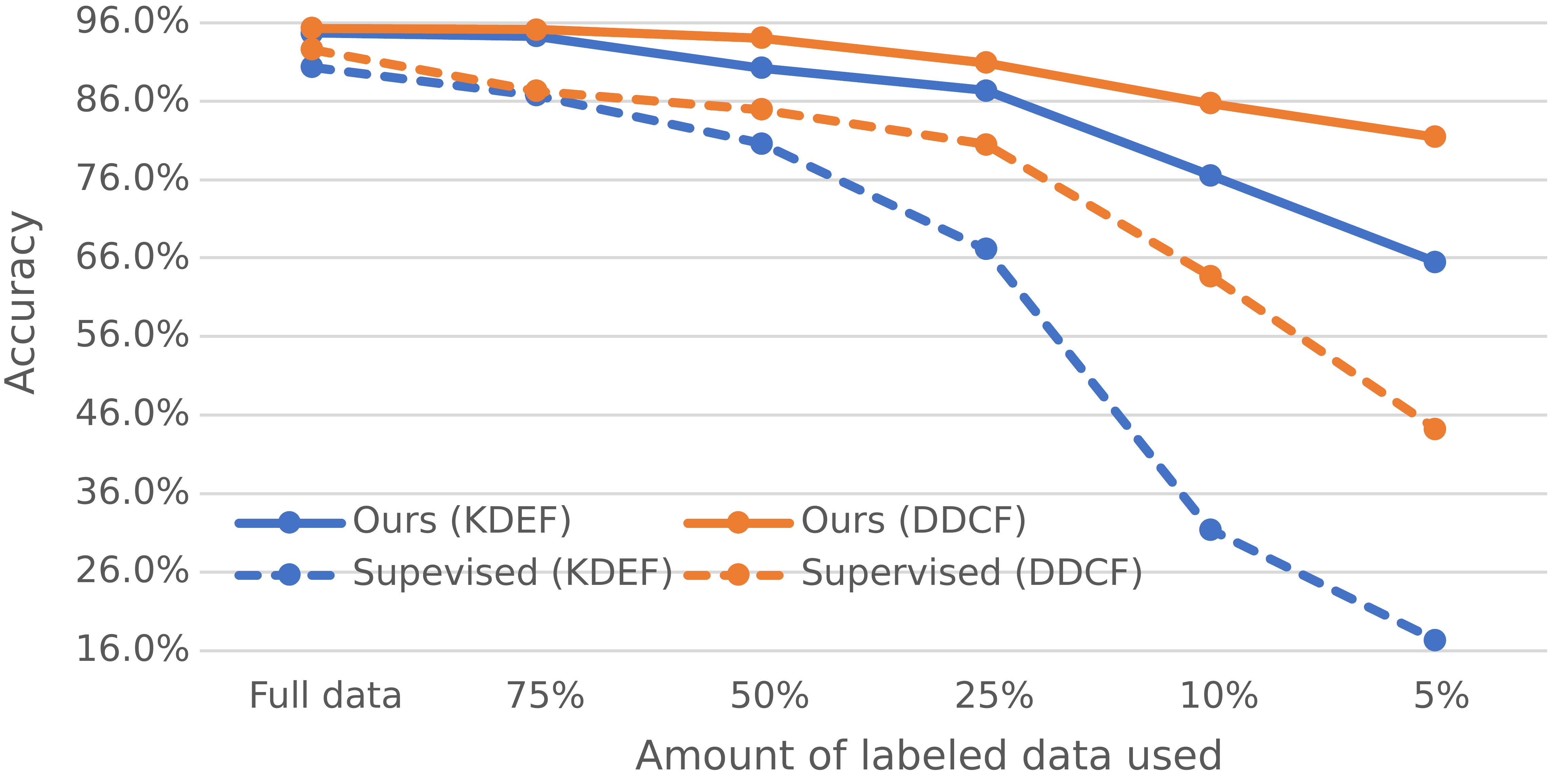}
  \caption{Performance of self-supervised CL-MEx and its fully supervised counterpart when different amounts of labeled data are used.}
  \label{fig:partial}
\end{figure}

Next, we perform a sensitivity study on the proposed method for the amount of labeled data by comparing the performance of CL-MEx to its fully supervised counterpart when 75\%, 50\%, 25\%, 10\%, and 5\% of the labels are used. The results are presented in Figure \ref{fig:partial}, where when all the data are used, our method shows 4\% and 3\% improvements over the fully supervised model for KDEF and DDCF datasets respectively. It is then seen that as the amount of output training labels are reduced, our method experiences smaller drops in performance versus the fully supervised model, showing a clear advantage by reduced reliance on the output labels. 


\section{Conclusion and Future Work}
This paper presents a self-supervised contrastive framework specialized in learning facial expressions from multiple views. Our method uses the multi-view images of the subjects during the pre-training and learns view-invariant embedding representations. The model is then fine-tuned in a supervised setting. We test our solution on KDEF and DDCF dataset and perform rigorous experiments. The model outperforms existing methods and shows robustness towards challenging viewing angle. Experiments also show the robustness of our model towards reduced amounts of labeled data.
For future work, the proposed method can be used for other multi-view or multi-modal image-based applications in self-supervised and contrastive learning settings, for example multi-view face/activity recognition or even multi-view object recognition.

\section*{Acknowledgements}
We would like to thank BMO Bank of Montreal and Mitacs for funding this research.

\bibliographystyle{ACM-Reference-Format}
\bibliography{references}

\end{document}